\documentclass[sigconf]{acmart}
\usepackage[caption=false]{subfig}
\usepackage[export]{adjustbox}
\usepackage{makecell}
\usepackage{amsmath,amsfonts}
\usepackage{algorithmic}
\usepackage{graphicx}
\usepackage{textcomp}
\usepackage{xcolor}
\usepackage{multirow}
\usepackage{hyperref}
\usepackage{booktabs}
\usepackage{hhline}
\usepackage[shortlabels]{enumitem}
\usepackage{soul}
\usepackage{fancyhdr}   
\usepackage{calc}
\usepackage{atbegshi}
\usepackage{tikz}

\usepackage{geometry}
\usepackage[utf8]{inputenc}

\AtBeginDocument{%
  }

\definecolor{rubinered}{HTML}{CE0058}

\definecolor{green}{rgb}{0.0, 0.65, 0.31}

\definecolor{bleudefrance}{rgb}{0.19, 0.55, 0.91}

\definecolor{ao(english)}{rgb}{0.0, 0.5, 0.0}

\definecolor{violet}{HTML}{6a51a3}

\begin{document}

\title{On the Benefit of Generative Foundation Models for Human Activity Recognition }

\author{Zikang Leng}
\email{zleng7@gatech.edu}
\orcid{0000-0001-6789-4780}
\affiliation{%
  \institution{Georgia Institute of Technology}
  \city{Atlanta, Georgia}
  \country{USA}}
  
\author{Hyeokhyen Kwon}
\email{hyeokhyen.kwon@emory.edu}
\orcid{0000-0002-5693-3278}
\affiliation{%
  \institution{Emory University}
  \city{Atlanta, Georgia}
  \country{USA}}
  
\author{Thomas Plötz}
\email{thomas.ploetz@gatech.edu}
\orcid{0000-0002-1243-7563}
\affiliation{%
  \institution{Georgia Institute of Technology}
  \city{Atlanta, Georgia}
  \country{USA}}

\renewcommand{\shortauthors}{Leng et al.}

\begin{abstract}
\noindent
In human activity recognition (HAR), the limited availability of annotated data presents a significant challenge. Drawing inspiration from the latest advancements in generative AI, including Large Language Models (LLMs) and motion synthesis models, we believe that generative AI can address this data scarcity by autonomously generating virtual IMU data from text descriptions. Beyond this, we spotlight several promising research pathways that could benefit from generative AI for the community, including the generating benchmark datasets, the development of foundational models specific to HAR, the exploration of hierarchical structures within HAR, breaking down complex activities, and applications in health sensing and activity summarization.


\vspace{-0.1in}
\end{abstract}

\ccsdesc[500]{Human-centered computing~Ubiquitous and mobile computing}
\ccsdesc[500]{Computing methodologies~Artificial intelligence}

\keywords{Virtual IMU Data; Activity recognition; Motion Synthesis; Large Language Models; Wearable Sensors}


\maketitle

\vspace{-0.1in}
\section{Introduction}
\noindent
Developing predictive models for Human Activity Recognition (HAR) that are both accurate and robust is important in various fields, such as fitness monitoring, health behavior analysis, and the optimization of industrial processes \cite{bachlin2010wearable, chavarriaga2013opportunity, Liaqat2019, Stiefmeier2008, Plötz2018mobile}.

The success of supervised learning techniques in developing HAR systems for wearable devices hinges largely on the presence of well-organized, annotated datasets \cite{chen2021sensecollect}. A key hurdle facing contemporary machine learning approaches in this area is the scarcity of labeled datasets \cite{Plötz2023data}. In HAR, labeling sensor data is not only costly but also frequently raises privacy concerns, and is susceptible to inaccuracies or other practical constraints \cite{kwon2019handling,jiang2021research,cilliers2020wearable}.


To address the issue of data scarcity, several studies have investigated using videos for training wearable-based HAR system \cite{Rey2019letIMU, Santhalingam2023asl, kwon2020imutube}, termed cross-modality transfer. Most notably, IMUTube \cite{kwon2020imutube, kwon2021approaching, kwon2021complex} extracts virtual IMU data from RGB human activity videos and has been shown to significantly improve the HAR model performance. However, IMUTube requires extensive computational power and often misses subtle movements in daily activities \cite{leng2022finegrained}.

With the rise of large language models (LLMs) like ChatGPT~\cite{brown2020gpt3}, Leng \textit{et al.}\cite{leng2023generating} combined LLMs with text-driven motion synthesis models to automatically generate virtual IMU data. Unlike traditional cross-modality methods, this approach harnesses the generative capabilities of foundational models to produce virtual IMU data. It eliminates the need for videos and delivers varied virtual IMU data for a wide range of real-world activity contexts.

Inspired by the results, we envision the rise of generative AI can lead to the generation of large-scale labeled training datasets to facilitate the training of complex and robust HAR models. Building on this vision, we suggest several future research trajectories for the community to consider:

\begin{enumerate}
[-,topsep=0pt]
\item Create expansive benchmark datasets for robust HAR model training and pretraining in self-supervised learning (SSL)

\item Conduct investigations into the hierarchical structures of human activities  and breaking down complex activities into simpler movements

\item Broader HAR applications in health sensing and activity summarization
\end{enumerate}

\vspace{-0.1in}



\vspace{-0.05in}
\section{Generating virtual IMU data using Generative AI}
\begin{figure}
    
    \centering
    \includegraphics[width=1\linewidth]{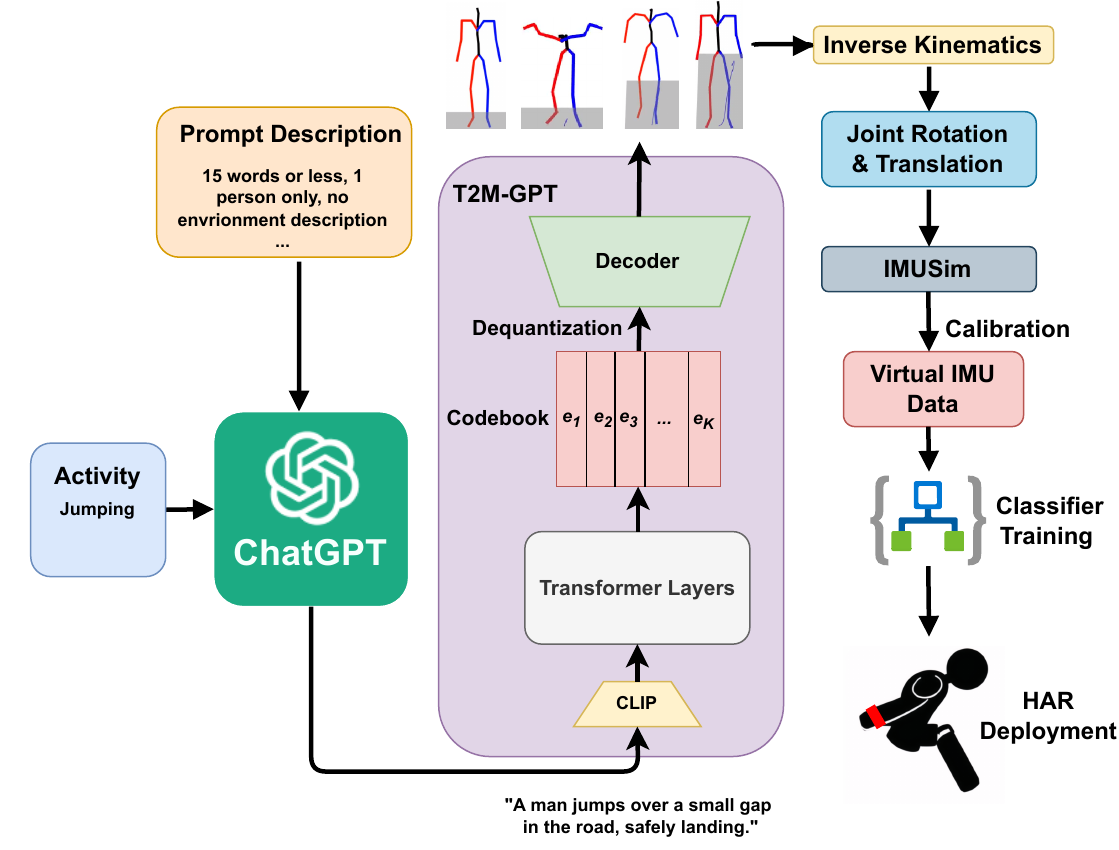}
     \vspace*{-0.3in}
    \caption{Overview of Leng \textit{et al.}'s approach \cite{leng2023generating}. 
    ChatGPT generates textual descriptions of activities. 
    T2M-GPT then generates 3D human motion sequences using the generated textual descriptions. 
    The motion sequences are then converted to virtual IMU data using inverse kinematics and IMUSim.    (Figure adopted from Leng et al. \cite{leng2023generating} and used with permission).
    }
    \label{fig:t2imu}
    \vspace*{-0.05in}
\end{figure}

\noindent
Recently, Leng \textit{et al.}~\cite{leng2023generating} introduced a pipeline capable of generating diverse textual descriptions of activities, which can then be transformed into virtual IMU data streams. This is achieved through the integration of ChatGPT, a motion synthesis model, and signal processing techniques, as shown in \autoref{fig:t2imu}.

The process begins with the provision of an activity name and general description of desired prompts. The pipeline then produces diverse and detailed textual depictions of a person engaging in the specified activity. These textual descriptions serve as prompts for the T2M-GPT motion synthesis model \cite{zhang2023generating}, which generates 3D human motion sequences. The 3D human motion can then be transformed into virtual IMU data streams via inverse kinematics and IMUSim. The virtual IMU data is then calibrated with a small amount of target real IMU data to to bridge the gap between two domains. The calibrated virtual IMU data can be used alongside real IMU data to train deployable HAR models.


This approach is inspired by the inherent variability in human activities. For example, a person may walk happily, confidently, quickly, or exhibit other variations in their movements. This diversity in human behavior is mirrored in the IMU data captured by wearable sensors. In order to ensure that HAR models can generalize well, it is important that the training data faithfully represents this variability. By leveraging ChatGPT to produce a wide range of textual descriptions of how activities can be executed (refer to \autoref{table:prompts}), this pipeline aims to generate virtual IMU data that encompasses the diversity present in real-world human activities. Such a variable virtual IMU dataset is expected to facilitate the training of HAR models that are both robust and generalizable. The effectiveness of this approach has been verified on three widely recognized HAR datasets - RealWorld \cite{Sztyler2016realworld}, Pamap2 \cite{Reisspamap}, and USC-HAD \cite{Zhang2012usc} - showing significant improvement in the performance of the downstream classifier.

Compared to video-based systems like IMUTube, this system offer a more streamlined process by eliminating the time-consuming task of video selection. 
Selecting suitable videos can be a labor-intensive process. By utilizing ChatGPT to automatically generate textual descriptions of activities, we can eliminate the need for manual effort. 
In essence, the generative AI-driven cross-modality approach offers a promising solution to the challenges of scale in labeled HAR datasets, achieving this cost-effectively and with minimal limitations on activity diversity. Given the advancements in LLMs and text-driven motion synthesis models, we anticipate the creation of robust HAR models. These models could encompass a broad spectrum of activities, including those difficult to gather due to privacy issues, practical limitations, or specific movement traits linked to various health conditions~\cite{Uhlenberg2023gait}.


\begin{table}
\centering
\caption{Example ChatGPT generated prompts for walking}
\vspace{-0.1in}
\begin{tabular}{|>{\footnotesize }p{0.64\columnwidth}|}
\hline
A woman strolls leisurely along the beach. \\
\hline
A man strides confidently through the city streets. \\
\hline
A woman shuffles tentatively along the icy sidewalk.\\
\hline
A woman strides purposefully towards the front door. \\
\hline
A man walks hesitantly towards the edge of a cliff. \\
\hline
A woman marches determinedly towards the finish line. \\
\hline
A retiree takes a long, therapeutic walk in the park. \\
\hline
A woman marches determinedly towards the finish line. \\
\hline
\end{tabular}


\label{table:prompts}
\end{table}

\vspace{-0.1in}
\section{The Way Forward}

In what follows, we outline promising future research directions that could transform the paradigm of HAR development with the emergence of generative AI.

\noindent\textbf{Generating Benchmark Datasets}
As shown in the previous section, generative AI is capable of generating diverse virtual IMU data. This paves the way for generating expansive virtual IMU datasets, encompassing wide range of activities in the real world
~\cite{nair2023dataset}. Distributing such comprehensive benchmark datasets could fill a significant gap in current practice for evaluating both performance and biases in HAR models. It offers a standardized dataset for comparative HAR model evaluation, addressing the current scenario where different studies utilize varied small benchmark datasets~\cite{Plötz2023data}.

\noindent\textbf{SSL in HAR}
The creation of a large-scale virtual IMU dataset can pave the way for the development complex HAR models using SSL techniques~\cite{Haresamudram2022ssl}. 
While recent strides in foundational models with SSL have shown remarkable few-shot or zero-shot capabilities in fields like natural language processing~\cite{brown2020gpt3} and computer vision~\cite{kirillov2023segany}, similar successes in SSL for HAR have been somewhat limited. These studies often rely on multiple smaller datasets, constraining our grasp of SSL's potential for robust representation learning. We anticipate that breakthroughs in generative AI will facilitate the generation of expansive HAR datasets, akin to those in other domains~\cite{Deng2009imagenet}. This, in turn, could catalyze the evolution of foundational HAR models, pushing the boundaries of HAR applications.

\noindent\textbf{Breaking Down Complex Activities} Athanasiou \textit{et al.} \cite{SINC:ICCV:2022} highlighted that motion information are encoded within LLMs and demonstrated the utility of GPT-3 \cite{brown2020gpt3} to identify body parts involved in specific actions. Given these findings, we advocate for research exploring the potential of LLMs to spatially decompose complex activities into their elementary actions. For instance, the "bench press" activity might be conceptualized as a synthesis of the actions "lying down" and "pushing arm up".

Beyond spatial decomposition, temporal decomposition of activities presents another avenue of exploration. The 3D human motion sequences produced by T2M-GPT are essentially a series of decoded latent vectors, derived from a codebook encompassing 512 entries. This suggests that any human motion sequence within the HumanML3D dataset can be understood as an combination of briefer motion sequences selected from this set of 512. From this perspective, human motion can be analogized to a language, wherein the codebook latent vectors represent its lexicon. An intriguing possibility lies in reversing this process: rather than generating human motion sequences, existing sequences could be deconstructed into their constituent latent vectors, potentially uncovering a 'grammar' of human motion.

Merging these perspectives allows for the deconstruction of complex activities into both their constituent body movements and their sequential elements. When paired with methods for converting 3D poses to IMU data, such insights can be further translated to sensor-based data representations.

\noindent\textbf{Activity Summarization} With LLMs, activity summarization becomes more accessible and efficient. Imagine someone wearing a smartwatch equipped with a basic HAR classifier. At the end of the day, the classifier's predictions can be processed by the LLM to produce a concise summary of the individual's activities. This not only helps in recollecting the significant events of the day but also in planning for the next. Moreover, such activity summarization has potential applications in elderly care, offering an alternative means to monitor daily routines of seniors who might have difficulty recalling their day-to-day activities.

\noindent\textbf{Hierarchical Structures in HAR}
Human activity inherently follows a hierarchical structure~\cite{Mahsun2021hiearchy}. Consider morning routines: they encompass a range of tasks like washing, grooming, and eating, which can be broken down into specific actions, such as washing dishes or making salads.
However, amassing a large-scale dataset of daily activities with detailed hierarchical annotations is a daunting task, primarily due to the need for manual annotations. 
LLMs can be used to generate descriptions of both high-level routines and their corresponding granular actions, which could serve as hierarchical annotations for the generated virtual IMU data. The hierarchical dataset can be used to train a versatile HAR model capable of detecting and passively monitoring a spectrum of activity levels. Moreover, the outputs from such a HAR model can be integrated with an LLM. This combined system can offer users activity summaries, personalized advice, or even long-term behavioral analyses. 
This combination can greatly enhance the capabilities of intelligent assistant systems paired with wearable sensors.

\noindent\textbf{Health Sensing Applications}
An important application avenue of HAR is in health sensing. The biomedical research sector is actively exploring using  wearable IMU sensors to measure neurological disorders like Parkinson's Disease~\cite{Buckley2019parkinson} and Stroke~\cite{Felius2022stroke} through movement analysis. However, these studies also face data scarcity issues as recruitment and annotation are extremely costly. 
The high costs associated with patient recruitment and dataset annotation are significant barriers. It's challenging to recruit patients due to the severe daily life challenges they face. Moreover, disorders like Rett Syndrome~\cite{Kyle2018rett} are so rare that finding suitable patients becomes even more difficult. Dataset annotation demands expertise, as specialists need to be well-acquainted with each specific disease condition. 
We believe that Biomedical LLMs~\cite{singhal2022medical} and motion synthesis models can offer a solution. They have the potential to sidestep data collection hurdles by supplying vast virtual IMU benchmark datasets with labels, which can then be used to train HAR models for clinical applications. 
To realize this vision, a collaborative effort between the HAR and Biomedical sectors is essential.

\vspace{-0.1in}
\section{Conclusion}

In this position paper, we explored the potential of recent breakthroughs in generative AI, especially LLMs and motion synthesis models, to produce extensive virtual IMU datasets. This approach can address the prevalent issue of insufficient annotated data within the HAR community. We envision that generative AI will play a pivotal role in advancing the development of robust HAR models and their applications, which include creating comprehensive HAR datasets, revealing the intricate hierarchical structures inherent in human activities, and tackling challenges in health sensing. 

\bibliographystyle{ACM-Reference-Format}
\bibliography{./bibs/virtual_imu,./bibs/motion_synthesis, ./bibs/har, ./bibs/llm}


\begin{thebibliography}{33}


\ifx \showCODEN    \undefined \def \showCODEN     #1{\unskip}     \fi
\ifx \showDOI      \undefined \def \showDOI       #1{#1}\fi
\ifx \showISBNx    \undefined \def \showISBNx     #1{\unskip}     \fi
\ifx \showISBNxiii \undefined \def \showISBNxiii  #1{\unskip}     \fi
\ifx \showISSN     \undefined \def \showISSN      #1{\unskip}     \fi
\ifx \showLCCN     \undefined \def \showLCCN      #1{\unskip}     \fi
\ifx \shownote     \undefined \def \shownote      #1{#1}          \fi
\ifx \showarticletitle \undefined \def \showarticletitle #1{#1}   \fi
\ifx \showURL      \undefined \def \showURL       {\relax}        \fi
\providecommand\bibfield[2]{#2}
\providecommand\bibinfo[2]{#2}
\providecommand\natexlab[1]{#1}
\providecommand\showeprint[2][]{arXiv:#2}

\bibitem[Altin et~al\mbox{.}(2021)]%
        {Mahsun2021hiearchy}
\bibfield{author}{\bibinfo{person}{Mahsun Altin}, \bibinfo{person}{Furkan
  Gursoy}, {and} \bibinfo{person}{Lina Xu}.} \bibinfo{year}{2021}\natexlab{}.
\newblock \showarticletitle{Machine-Generated Hierarchical Structure of Human
  Activities to Reveal How Machines Think}.
\newblock \bibinfo{journal}{\emph{{IEEE} Access}}  \bibinfo{volume}{9}
  (\bibinfo{year}{2021}), \bibinfo{pages}{18307--18317}.
\newblock
\urldef\tempurl%
\url{https://doi.org/10.1109/access.2021.3053084}
\showDOI{\tempurl}


\bibitem[Athanasiou et~al\mbox{.}(2023)]%
        {SINC:ICCV:2022}
\bibfield{author}{\bibinfo{person}{Nikos Athanasiou}, \bibinfo{person}{Mathis
  Petrovich}, \bibinfo{person}{Michael~J. Black}, {and}
  \bibinfo{person}{G\"{u}l Varol}.} \bibinfo{year}{2023}\natexlab{}.
\newblock \showarticletitle{{SINC}: Spatial Composition of {3D} Human Motions
  for Simultaneous Action Generation}. In \bibinfo{booktitle}{\emph{ICCV}}.
\newblock


\bibitem[B{\"a}chlin et~al\mbox{.}(2010)]%
        {bachlin2010wearable}
\bibfield{author}{\bibinfo{person}{Matthias B{\"a}chlin}, \bibinfo{person}{Meir
  Plotnik}, \bibinfo{person}{Daniel Roggen}, \bibinfo{person}{Nir Giladi},
  \bibinfo{person}{Jeffrey~M Hausdorff}, {and} \bibinfo{person}{Gerhard
  Tr{\"o}ster}.} \bibinfo{year}{2010}\natexlab{}.
\newblock \showarticletitle{Wearable assistant for Parkinson's disease patients
  with the freezing of gait symptom}.
\newblock \bibinfo{journal}{\emph{IEEE Transactions on Information Technology
  in Biomedicine}} \bibinfo{volume}{14}, \bibinfo{number}{2}
  (\bibinfo{year}{2010}), \bibinfo{pages}{436--446}.
\newblock
\urldef\tempurl%
\url{https://doi.org/10.1109/TITB.2009.2036165}
\showDOI{\tempurl}


\bibitem[Brown et~al\mbox{.}(2020)]%
        {brown2020gpt3}
\bibfield{author}{\bibinfo{person}{Tom Brown}, \bibinfo{person}{Benjamin Mann},
  \bibinfo{person}{Nick Ryder}, \bibinfo{person}{Melanie Subbiah},
  \bibinfo{person}{Jared~D Kaplan}, \bibinfo{person}{Prafulla Dhariwal},
  \bibinfo{person}{Arvind Neelakantan}, \bibinfo{person}{Pranav Shyam},
  \bibinfo{person}{Girish Sastry}, \bibinfo{person}{Amanda Askell},
  \bibinfo{person}{Sandhini Agarwal}, \bibinfo{person}{Ariel Herbert-Voss},
  \bibinfo{person}{Gretchen Krueger}, \bibinfo{person}{Tom Henighan},
  \bibinfo{person}{Rewon Child}, \bibinfo{person}{Aditya Ramesh},
  \bibinfo{person}{Daniel Ziegler}, \bibinfo{person}{Jeffrey Wu},
  \bibinfo{person}{Clemens Winter}, \bibinfo{person}{Chris Hesse},
  \bibinfo{person}{Mark Chen}, \bibinfo{person}{Eric Sigler},
  \bibinfo{person}{Mateusz Litwin}, \bibinfo{person}{Scott Gray},
  \bibinfo{person}{Benjamin Chess}, \bibinfo{person}{Jack Clark},
  \bibinfo{person}{Christopher Berner}, \bibinfo{person}{Sam McCandlish},
  \bibinfo{person}{Alec Radford}, \bibinfo{person}{Ilya Sutskever}, {and}
  \bibinfo{person}{Dario Amodei}.} \bibinfo{year}{2020}\natexlab{}.
\newblock \showarticletitle{Language Models are Few-Shot Learners}. In
  \bibinfo{booktitle}{\emph{Advances in Neural Information Processing
  Systems}}, Vol.~\bibinfo{volume}{33}. \bibinfo{publisher}{Curran Associates,
  Inc.}, \bibinfo{pages}{1877--1901}.
\newblock


\bibitem[Buckley et~al\mbox{.}(2019)]%
        {Buckley2019parkinson}
\bibfield{author}{\bibinfo{person}{Christopher Buckley}, \bibinfo{person}{Lisa
  Alcock}, \bibinfo{person}{Riona McArdle}, \bibinfo{person}{Rana~Rashid
  Rehman}, \bibinfo{person}{Silvia~Del Din}, \bibinfo{person}{Claudia
  Mazz{\`a}}, \bibinfo{person}{Alison~J. Yarnall}, {and} \bibinfo{person}{Lynn
  Rochester}.} \bibinfo{year}{2019}\natexlab{}.
\newblock \showarticletitle{The Role of Movement Analysis in Diagnosing and
  Monitoring Neurodegenerative Conditions: Insights from Gait and Postural
  Control}.
\newblock \bibinfo{journal}{\emph{Brain Sciences}}  \bibinfo{volume}{9}
  (\bibinfo{year}{2019}).
\newblock
\urldef\tempurl%
\url{https://api.semanticscholar.org/CorpusID:73425357}
\showURL{%
\tempurl}


\bibitem[Chavarriaga et~al\mbox{.}(2013)]%
        {chavarriaga2013opportunity}
\bibfield{author}{\bibinfo{person}{Ricardo Chavarriaga}, \bibinfo{person}{Hesam
  Sagha}, \bibinfo{person}{Alberto Calatroni}, \bibinfo{person}{Sundara~Tejaswi
  Digumarti}, \bibinfo{person}{Gerhard Tr{\"o}ster}, \bibinfo{person}{Jos{\'e}
  del~R Mill{\'a}n}, {and} \bibinfo{person}{Daniel Roggen}.}
  \bibinfo{year}{2013}\natexlab{}.
\newblock \showarticletitle{The opportunity challenge: a benchmark database for
  on-body sensor-based activity recognition}.
\newblock \bibinfo{journal}{\emph{Pattern Recognition Letters}}
  \bibinfo{volume}{34}, \bibinfo{number}{15} (\bibinfo{year}{2013}),
  \bibinfo{pages}{2033--2042}.
\newblock
\urldef\tempurl%
\url{https://doi.org/10.1016/j.patrec.2012.12.014}
\showDOI{\tempurl}


\bibitem[Chen et~al\mbox{.}(2021)]%
        {chen2021sensecollect}
\bibfield{author}{\bibinfo{person}{Wenqiang Chen}, \bibinfo{person}{Shupei
  Lin}, \bibinfo{person}{Elizabeth Thompson}, {and} \bibinfo{person}{John
  Stankovic}.} \bibinfo{year}{2021}\natexlab{}.
\newblock \showarticletitle{SenseCollect: We Need Efficient Ways to Collect
  On-body Sensor-based Human Activity Data!}
\newblock \bibinfo{journal}{\emph{Proceedings of the ACM on Interactive,
  Mobile, Wearable and Ubiquitous Technologies}} \bibinfo{volume}{5},
  \bibinfo{number}{3} (\bibinfo{year}{2021}), \bibinfo{pages}{1--27}.
\newblock


\bibitem[Cilliers(2020)]%
        {cilliers2020wearable}
\bibfield{author}{\bibinfo{person}{L. Cilliers}.}
  \bibinfo{year}{2020}\natexlab{}.
\newblock \showarticletitle{Wearable devices in healthcare: Privacy and
  information security issues}.
\newblock \bibinfo{journal}{\emph{Health information management journal}}
  \bibinfo{volume}{49}, \bibinfo{number}{2-3} (\bibinfo{year}{2020}),
  \bibinfo{pages}{150--156}.
\newblock


\bibitem[Deng et~al\mbox{.}(2009)]%
        {Deng2009imagenet}
\bibfield{author}{\bibinfo{person}{Jia Deng}, \bibinfo{person}{Wei Dong},
  \bibinfo{person}{Richard Socher}, \bibinfo{person}{Li-Jia Li},
  \bibinfo{person}{Kai Li}, {and} \bibinfo{person}{Li Fei-Fei}.}
  \bibinfo{year}{2009}\natexlab{}.
\newblock \showarticletitle{ImageNet: A large-scale hierarchical image
  database}. In \bibinfo{booktitle}{\emph{2009 IEEE Conference on Computer
  Vision and Pattern Recognition}}. \bibinfo{pages}{248--255}.
\newblock
\urldef\tempurl%
\url{https://doi.org/10.1109/CVPR.2009.5206848}
\showDOI{\tempurl}


\bibitem[Felius et~al\mbox{.}(2022)]%
        {Felius2022stroke}
\bibfield{author}{\bibinfo{person}{R.A.W. Felius}, \bibinfo{person}{M.
  Geerars}, \bibinfo{person}{S.M. Bruijn}, \bibinfo{person}{N.C. Wouda},
  \bibinfo{person}{J.H. {Van Dieën}}, {and} \bibinfo{person}{M. Punt}.}
  \bibinfo{year}{2022}\natexlab{}.
\newblock \showarticletitle{Reliability of IMU-based balance assessment in
  clinical stroke rehabilitation}.
\newblock \bibinfo{journal}{\emph{Sensors}} (\bibinfo{year}{2022}).
\newblock


\bibitem[Haresamudram et~al\mbox{.}(2022)]%
        {Haresamudram2022ssl}
\bibfield{author}{\bibinfo{person}{Harish Haresamudram}, \bibinfo{person}{Irfan
  Essa}, {and} \bibinfo{person}{Thomas Pl\"{o}tz}.}
  \bibinfo{year}{2022}\natexlab{}.
\newblock \showarticletitle{Assessing the State of Self-Supervised Human
  Activity Recognition Using Wearables}.
\newblock \bibinfo{journal}{\emph{Proc. ACM Interact. Mob. Wearable Ubiquitous
  Technol.}} (\bibinfo{year}{2022}).
\newblock


\bibitem[Jiang and Shi(2021)]%
        {jiang2021research}
\bibfield{author}{\bibinfo{person}{D. Jiang} {and} \bibinfo{person}{G. Shi}.}
  \bibinfo{year}{2021}\natexlab{}.
\newblock \showarticletitle{Research on data security and privacy protection of
  wearable equipment in healthcare}.
\newblock \bibinfo{journal}{\emph{Journal of Healthcare Engineering}}
  \bibinfo{volume}{2021} (\bibinfo{year}{2021}).
\newblock


\bibitem[Kirillov et~al\mbox{.}(2023)]%
        {kirillov2023segany}
\bibfield{author}{\bibinfo{person}{Alexander Kirillov}, \bibinfo{person}{Eric
  Mintun}, \bibinfo{person}{Nikhila Ravi}, \bibinfo{person}{Hanzi Mao},
  \bibinfo{person}{Chloe Rolland}, \bibinfo{person}{Laura Gustafson},
  \bibinfo{person}{Tete Xiao}, \bibinfo{person}{Spencer Whitehead},
  \bibinfo{person}{Alexander~C. Berg}, \bibinfo{person}{Wan-Yen Lo},
  \bibinfo{person}{Piotr Doll{\'a}r}, {and} \bibinfo{person}{Ross Girshick}.}
  \bibinfo{year}{2023}\natexlab{}.
\newblock \showarticletitle{Segment Anything}.
\newblock \bibinfo{journal}{\emph{arXiv:2304.02643}} (\bibinfo{year}{2023}).
\newblock


\bibitem[Kwon et~al\mbox{.}(2019)]%
        {kwon2019handling}
\bibfield{author}{\bibinfo{person}{Hyeokhyen Kwon}, \bibinfo{person}{Gregory~D
  Abowd}, {and} \bibinfo{person}{Thomas Pl{\"o}tz}.}
  \bibinfo{year}{2019}\natexlab{}.
\newblock \showarticletitle{Handling annotation uncertainty in human activity
  recognition}. In \bibinfo{booktitle}{\emph{Proceedings of the 23rd
  International Symposium on Wearable Computers}}. \bibinfo{pages}{109--117}.
\newblock


\bibitem[Kwon et~al\mbox{.}(2021a)]%
        {kwon2021complex}
\bibfield{author}{\bibinfo{person}{Hyeokhyen Kwon}, \bibinfo{person}{Gregory~D
  Abowd}, {and} \bibinfo{person}{Thomas Pl{\"o}tz}.}
  \bibinfo{year}{2021}\natexlab{a}.
\newblock \showarticletitle{Complex Deep Neural Networks from Large Scale
  Virtual IMU Data for Effective Human Activity Recognition Using Wearables}.
\newblock \bibinfo{journal}{\emph{Sensors}} \bibinfo{volume}{21},
  \bibinfo{number}{24} (\bibinfo{year}{2021}), \bibinfo{pages}{8337}.
\newblock


\bibitem[Kwon et~al\mbox{.}(2020)]%
        {kwon2020imutube}
\bibfield{author}{\bibinfo{person}{Hyeokhyen Kwon}, \bibinfo{person}{Catherine
  Tong}, \bibinfo{person}{Harish Haresamudram}, \bibinfo{person}{Yan Gao},
  \bibinfo{person}{Gregory~D Abowd}, \bibinfo{person}{Nicholas~D Lane}, {and}
  \bibinfo{person}{Thomas Ploetz}.} \bibinfo{year}{2020}\natexlab{}.
\newblock \showarticletitle{Imutube: Automatic extraction of virtual on-body
  accelerometry from video for human activity recognition}.
\newblock \bibinfo{journal}{\emph{Proceedings of the ACM on Interactive,
  Mobile, Wearable and Ubiquitous Technologies}} \bibinfo{volume}{4},
  \bibinfo{number}{3} (\bibinfo{year}{2020}), \bibinfo{pages}{1--29}.
\newblock


\bibitem[Kwon et~al\mbox{.}(2021b)]%
        {kwon2021approaching}
\bibfield{author}{\bibinfo{person}{Hyeokhyen Kwon}, \bibinfo{person}{Bingyao
  Wang}, \bibinfo{person}{Gregory~D Abowd}, {and} \bibinfo{person}{Thomas
  Pl{\"o}tz}.} \bibinfo{year}{2021}\natexlab{b}.
\newblock \showarticletitle{Approaching the Real-World: Supporting Activity
  Recognition Training with Virtual IMU Data}.
\newblock \bibinfo{journal}{\emph{Proceedings of the ACM on Interactive,
  Mobile, Wearable and Ubiquitous Technologies}} \bibinfo{volume}{5},
  \bibinfo{number}{3} (\bibinfo{year}{2021}), \bibinfo{pages}{1--32}.
\newblock


\bibitem[Kyle et~al\mbox{.}(2018)]%
        {Kyle2018rett}
\bibfield{author}{\bibinfo{person}{Stephanie Kyle}, \bibinfo{person}{Neeti
  Vashi}, {and} \bibinfo{person}{Monica Justice}.}
  \bibinfo{year}{2018}\natexlab{}.
\newblock \showarticletitle{Rett syndrome: A neurological disorder with
  metabolic components}.
\newblock \bibinfo{journal}{\emph{Open Biology}}  \bibinfo{volume}{8}
  (\bibinfo{date}{02} \bibinfo{year}{2018}), \bibinfo{pages}{170216}.
\newblock
\urldef\tempurl%
\url{https://doi.org/10.1098/rsob.170216}
\showDOI{\tempurl}


\bibitem[Leng et~al\mbox{.}(2022)]%
        {leng2022finegrained}
\bibfield{author}{\bibinfo{person}{Zikang Leng}, \bibinfo{person}{Yash Jain},
  \bibinfo{person}{Hyeokhyen Kwon}, {and} \bibinfo{person}{Thomas Plötz}.}
  \bibinfo{year}{2022}\natexlab{}.
\newblock \bibinfo{title}{Fine-grained Human Activity Recognition Using Virtual
  On-body Acceleration Data}.
\newblock
\newblock
\showeprint[arxiv]{2211.01342}~[cs.CV]


\bibitem[Leng et~al\mbox{.}(2023)]%
        {leng2023generating}
\bibfield{author}{\bibinfo{person}{Zikang Leng}, \bibinfo{person}{Hyeokhyen
  Kwon}, {and} \bibinfo{person}{Thomas Plötz}.}
  \bibinfo{year}{2023}\natexlab{}.
\newblock \bibinfo{title}{Generating Virtual On-body Accelerometer Data from
  Virtual Textual Descriptions for Human Activity Recognition}.
\newblock
\newblock
\showeprint[arxiv]{2305.03187}~[cs.CV]


\bibitem[Liaqat et~al\mbox{.}(2019)]%
        {Liaqat2019}
\bibfield{author}{\bibinfo{person}{Daniyal Liaqat}, \bibinfo{person}{Mohamed
  Abdalla}, \bibinfo{person}{Pegah Abed-Esfahani}, \bibinfo{person}{Moshe
  Gabel}, \bibinfo{person}{Tatiana Son}, \bibinfo{person}{Robert Wu},
  \bibinfo{person}{Andrea Gershon}, \bibinfo{person}{Frank Rudzicz}, {and}
  \bibinfo{person}{Eyal~De Lara}.} \bibinfo{year}{2019}\natexlab{}.
\newblock \showarticletitle{WearBreathing: Real World Respiratory Rate
  Monitoring Using Smartwatches}.
\newblock \bibinfo{journal}{\emph{Proc. ACM Interact. Mob. Wearable Ubiquitous
  Technol.}} \bibinfo{volume}{3}, \bibinfo{number}{2}, Article
  \bibinfo{articleno}{56} (\bibinfo{date}{jun} \bibinfo{year}{2019}),
  \bibinfo{numpages}{22}~pages.
\newblock
\urldef\tempurl%
\url{https://doi.org/10.1145/3328927}
\showDOI{\tempurl}


\bibitem[Nair et~al\mbox{.}(2023)]%
        {nair2023dataset}
\bibfield{author}{\bibinfo{person}{Nilah~Ravi Nair}, \bibinfo{person}{Lena
  Schmid}, \bibinfo{person}{Fernando~Moya Rueda}, \bibinfo{person}{Markus
  Pauly}, \bibinfo{person}{Gernot~A. Fink}, {and} \bibinfo{person}{Christopher
  Reining}.} \bibinfo{year}{2023}\natexlab{}.
\newblock \bibinfo{title}{Dataset Bias in Human Activity Recognition}.
\newblock
\newblock
\showeprint[arxiv]{2301.10161}~[eess.SP]


\bibitem[Plötz(2023)]%
        {Plötz2023data}
\bibfield{author}{\bibinfo{person}{Thomas Plötz}.}
  \bibinfo{year}{2023}\natexlab{}.
\newblock \showarticletitle{If only we had more data!: Sensor-Based Human
  Activity Recognition in Challenging Scenarios}. In
  \bibinfo{booktitle}{\emph{2023 IEEE International Conference on Pervasive
  Computing and Communications Workshops and other Affiliated Events (PerCom
  Workshops)}}. \bibinfo{pages}{565--570}.
\newblock
\urldef\tempurl%
\url{https://doi.org/10.1109/PerComWorkshops56833.2023.10150267}
\showDOI{\tempurl}


\bibitem[Plötz and Guan(2018)]%
        {Plötz2018mobile}
\bibfield{author}{\bibinfo{person}{Thomas Plötz} {and} \bibinfo{person}{Yu
  Guan}.} \bibinfo{year}{2018}\natexlab{}.
\newblock \showarticletitle{Deep Learning for Human Activity Recognition in
  Mobile Computing}.
\newblock \bibinfo{journal}{\emph{Computer}} \bibinfo{volume}{51},
  \bibinfo{number}{5} (\bibinfo{year}{2018}), \bibinfo{pages}{50--59}.
\newblock
\urldef\tempurl%
\url{https://doi.org/10.1109/MC.2018.2381112}
\showDOI{\tempurl}


\bibitem[Reiss and Stricker(2012)]%
        {Reisspamap}
\bibfield{author}{\bibinfo{person}{Attila Reiss} {and} \bibinfo{person}{Didier
  Stricker}.} \bibinfo{year}{2012}\natexlab{}.
\newblock \showarticletitle{Introducing a New Benchmarked Dataset for Activity
  Monitoring} \emph{(\bibinfo{series}{ISWC '12})}. \bibinfo{publisher}{IEEE
  Computer Society}.
\newblock
\showISBNx{9780769546971}
\urldef\tempurl%
\url{https://doi.org/10.1109/ISWC.2012.13}
\showDOI{\tempurl}


\bibitem[Rey et~al\mbox{.}(2019)]%
        {Rey2019letIMU}
\bibfield{author}{\bibinfo{person}{Vitor~Fortes Rey}, \bibinfo{person}{Peter
  Hevesi}, \bibinfo{person}{Onorina Kovalenko}, {and} \bibinfo{person}{Paul
  Lukowicz}.} \bibinfo{year}{2019}\natexlab{}.
\newblock \showarticletitle{Let There Be IMU Data: Generating Training Data for
  Wearable, Motion Sensor Based Activity Recognition from Monocular RGB Videos}
  \emph{(\bibinfo{series}{UbiComp/ISWC '19 Adjunct})}.
  \bibinfo{publisher}{Association for Computing Machinery}.
\newblock


\bibitem[Santhalingam et~al\mbox{.}(2023)]%
        {Santhalingam2023asl}
\bibfield{author}{\bibinfo{person}{Panneer~Selvam Santhalingam},
  \bibinfo{person}{Parth Pathak}, \bibinfo{person}{Huzefa Rangwala}, {and}
  \bibinfo{person}{Jana Kosecka}.} \bibinfo{year}{2023}\natexlab{}.
\newblock \showarticletitle{Synthetic Smartwatch IMU Data Generation from
  In-the-Wild ASL Videos}.
\newblock \bibinfo{journal}{\emph{Proc. ACM Interact. Mob. Wearable Ubiquitous
  Technol.}} (\bibinfo{year}{2023}).
\newblock


\bibitem[Singhal et~al\mbox{.}(2022)]%
        {singhal2022medical}
\bibfield{author}{\bibinfo{person}{Karan Singhal}, \bibinfo{person}{Shekoofeh
  Azizi}, \bibinfo{person}{Tao Tu}, \bibinfo{person}{S.~Sara Mahdavi},
  \bibinfo{person}{Jason Wei}, \bibinfo{person}{Hyung~Won Chung},
  \bibinfo{person}{Nathan Scales}, \bibinfo{person}{Ajay Tanwani},
  \bibinfo{person}{Heather Cole-Lewis}, \bibinfo{person}{Stephen Pfohl},
  \bibinfo{person}{Perry Payne}, \bibinfo{person}{Martin Seneviratne},
  \bibinfo{person}{Paul Gamble}, \bibinfo{person}{Chris Kelly},
  \bibinfo{person}{Nathaneal Scharli}, \bibinfo{person}{Aakanksha Chowdhery},
  \bibinfo{person}{Philip Mansfield}, \bibinfo{person}{Blaise~Aguera y Arcas},
  \bibinfo{person}{Dale Webster}, \bibinfo{person}{Greg~S. Corrado},
  \bibinfo{person}{Yossi Matias}, \bibinfo{person}{Katherine Chou},
  \bibinfo{person}{Juraj Gottweis}, \bibinfo{person}{Nenad Tomasev},
  \bibinfo{person}{Yun Liu}, \bibinfo{person}{Alvin Rajkomar},
  \bibinfo{person}{Joelle Barral}, \bibinfo{person}{Christopher Semturs},
  \bibinfo{person}{Alan Karthikesalingam}, {and} \bibinfo{person}{Vivek
  Natarajan}.} \bibinfo{year}{2022}\natexlab{}.
\newblock \bibinfo{title}{Large Language Models Encode Clinical Knowledge}.
\newblock
\newblock
\showeprint[arxiv]{2212.13138}~[cs.CL]


\bibitem[Stiefmeier et~al\mbox{.}(2008)]%
        {Stiefmeier2008}
\bibfield{author}{\bibinfo{person}{Thomas Stiefmeier}, \bibinfo{person}{Daniel
  Roggen}, \bibinfo{person}{Georg Ogris}, \bibinfo{person}{Paul Lukowicz},
  {and} \bibinfo{person}{Gerhard Tröster}.} \bibinfo{year}{2008}\natexlab{}.
\newblock \showarticletitle{Wearable Activity Tracking in Car Manufacturing}.
\newblock \bibinfo{journal}{\emph{IEEE Pervasive Computing}}
  \bibinfo{volume}{7}, \bibinfo{number}{2} (\bibinfo{year}{2008}),
  \bibinfo{pages}{42--50}.
\newblock
\urldef\tempurl%
\url{https://doi.org/10.1109/MPRV.2008.40}
\showDOI{\tempurl}


\bibitem[Sztyler and Stuckenschmidt(2016)]%
        {Sztyler2016realworld}
\bibfield{author}{\bibinfo{person}{Timo Sztyler} {and} \bibinfo{person}{Heiner
  Stuckenschmidt}.} \bibinfo{year}{2016}\natexlab{}.
\newblock \showarticletitle{On-body localization of wearable devices: An
  investigation of position-aware activity recognition}. In
  \bibinfo{booktitle}{\emph{2016 IEEE International Conference on Pervasive
  Computing and Communications (PerCom)}}. \bibinfo{pages}{1--9}.
\newblock
\urldef\tempurl%
\url{https://doi.org/10.1109/PERCOM.2016.7456521}
\showDOI{\tempurl}


\bibitem[Uhlenberg et~al\mbox{.}(2023)]%
        {Uhlenberg2023gait}
\bibfield{author}{\bibinfo{person}{Lena Uhlenberg}, \bibinfo{person}{Adrian
  Derungs}, {and} \bibinfo{person}{Oliver Amft}.}
  \bibinfo{year}{2023}\natexlab{}.
\newblock \showarticletitle{Co-simulation of human digital twins and wearable
  inertial sensors to analyse gait event estimation}.
\newblock \bibinfo{journal}{\emph{Frontiers in Bioengineering and
  Biotechnology}} (\bibinfo{year}{2023}).
\newblock


\bibitem[Zhang et~al\mbox{.}(2023)]%
        {zhang2023generating}
\bibfield{author}{\bibinfo{person}{Jianrong Zhang}, \bibinfo{person}{Yangsong
  Zhang}, \bibinfo{person}{Xiaodong Cun}, \bibinfo{person}{Shaoli Huang},
  \bibinfo{person}{Yong Zhang}, \bibinfo{person}{Hongwei Zhao},
  \bibinfo{person}{Hongtao Lu}, {and} \bibinfo{person}{Xi Shen}.}
  \bibinfo{year}{2023}\natexlab{}.
\newblock \showarticletitle{T2M-GPT: Generating Human Motion from Textual
  Descriptions with Discrete Representations}. In
  \bibinfo{booktitle}{\emph{Proceedings of the IEEE/CVF Conference on Computer
  Vision and Pattern Recognition (CVPR)}}.
\newblock


\bibitem[Zhang and Sawchuk(2012)]%
        {Zhang2012usc}
\bibfield{author}{\bibinfo{person}{Mi Zhang} {and}
  \bibinfo{person}{Alexander~A. Sawchuk}.} \bibinfo{year}{2012}\natexlab{}.
\newblock \showarticletitle{USC-HAD: A Daily Activity Dataset for Ubiquitous
  Activity Recognition Using Wearable Sensors}. \bibinfo{publisher}{Association
  for Computing Machinery}.
\newblock


\end{thebibliography}

\end{document}